\documentclass[sigconf]{acmart}

\usepackage{booktabs} % For formal tables
\usepackage{color}

\newcommand{\modelName}{RSRL{}} 
\newcommand{\modelNameFull}{Race Strategy Reinforcement Learning}

\copyrightyear{2025}
\acmYear{2025}
\setcopyright{cc}
\setcctype[4.0]{by}
\acmConference[SAC '25]{The 40th ACM/SIGAPP Symposium on Applied Computing}{March 31-April 4, 2025}{Catania, Italy}
\acmBooktitle{The 40th ACM/SIGAPP Symposium on Applied Computing (SAC '25), March 31-April 4, 2025, Catania, Italy}\acmDOI{10.1145/3672608.3707766}
\acmISBN{979-8-4007-0629-5/25/03}

\acmArticle{4}
\usepackage{savesym}
\savesymbol{Bbbk}
\savesymbol{openbox}

\usepackage{xcolor}
\usepackage{url}
\usepackage[utf8]{inputenc}
\usepackage{graphicx}
\usepackage{amsmath}
\usepackage{amssymb}
\usepackage{amsthm}
\usepackage{MnSymbol}
\usepackage{booktabs}
\usepackage{algorithm}
\usepackage{algorithmic}
\usepackage{svg}
\usepackage{longtable}
\usepackage{multirow}
\usepackage{subcaption}
\usepackage{afterpage}
\usepackage{tablefootnote}
\restoresymbol{NEW}{Bbbk}
\restoresymbol{NEW}{openbox}

\newcommand{\AR}[1]{\textcolor{black}{#1}}

\newcommand{\DT}[1]{\textcolor{black}{#1}}

\begin{document}
\title{Explainable Reinforcement Learning for \\ Formula One Race Strategy}
%\titlenote{Produces the permission block, and copyright information}
%\subtitle{Extended Abstract}
%\subtitlenote{The full version of the author's guide is available as \texttt{acmart.pdf} document}
\titlenote{© ACM 2025. This is the authors' version of the work. It is posted here for your personal use. Not for redistribution. The definitive Version of Record will be published in SAC 2025, \href{http://dx.doi.org/10.1145/3672608.3707766}{http://dx.doi.org/10.1145/3672608.3707766}.}

\renewcommand{\shorttitle}{SIG Proceedings Paper in LaTeX Format}

%\author{Anonymous Authors}

\author{Devin Thomas}
\orcid{1234-5678-9012}
\affiliation{
  \institution{Department of Computing, \\ Imperial College London, UK}
  \country{}
}
\email{devin.thomas20@imperial.ac.uk}

\author{Junqi Jiang}
\affiliation{
  \institution{Department of Computing, \\ Imperial College London, UK}
\country{}
}
\email{junqi.jiang20@imperial.ac.uk}

\author{Avinash Kori}
\affiliation{
  \institution{Department of Computing, \\ Imperial College London, UK}
\country{}
}
\email{a.kori21@imperial.ac.uk}

\author{Aaron Russo}
\affiliation{
    \institution{Mercedes-AMG PETRONAS F1 Team}
    \city{Brackley}
    \country{UK}
}
\email{arusso@mercedesamgf1.com}

\author{Steffen Winkler}
\affiliation{
    \institution{Mercedes-AMG PETRONAS F1 Team}
    \city{Brackley}
    \country{UK}
}
\email{swinkler@mercedesamgf1.com}

\author{Stuart Sale}
\affiliation{
    \institution{Mercedes-AMG PETRONAS F1 Team}
    \city{Brackley}
    \country{UK}
}
\email{ssale@mercedesamgf1.com}

\author{Joseph McMillan}
\affiliation{
    \institution{Mercedes-AMG PETRONAS F1 Team}
    \city{Brackley}
    \country{UK}
}
\email{jmcmillan@mercedesamgf1.com}

\author{Francesco Belardinelli}
\affiliation{
  \institution{Department of Computing, \\ Imperial College London, UK}
  \country{}
}
\email{francesco.belardinelli@imperial.ac.uk}

\author{Antonio Rago}
\authornote{Corresponding author.}
\affiliation{
  \institution{Department of Computing, \\ Imperial College London, UK}
  \country{}
}
\email{a.rago@imperial.ac.uk}

% The default list of authors is too long for headers}
\renewcommand{\shortauthors}{D. Thomas et al.}

\begin{abstract}
In Formula One, teams compete to develop their cars and achieve the highest possible finishing position in each race. During a race, however, teams are unable to alter the car, so they must improve their cars' finishing positions via race strategy, i.e. optimising their selection of which tyre compounds to put on the car and when to do so. In this work, we introduce a reinforcement learning model, \modelName{} (\modelNameFull{}%, pronounced `\modelPronunciation{}'
), to control race strategies in simulations, offering a faster alternative to the industry standard of hard-coded and Monte Carlo-based race strategies. Controlling cars with a pace equating to an expected finishing position of P5.5 (where P1 represents first place and P20 is last place), \modelName{} achieves an average finishing position of P5.33 on our test race, the 2023 Bahrain Grand Prix, outperforming the best baseline of P5.63. We then demonstrate, in a generalisability study, how performance for one track or multiple tracks can be prioritised via training. Further, we supplement model predictions with feature importance, decision tree-based surrogate models, and decision tree counterfactuals towards improving user trust in the model. Finally, we provide illustrations which exemplify our approach in real-world situations, drawing parallels between simulations and reality.
\end{abstract}

%
% The code below should be generated by the tool at
% http://dl.acm.org/ccs.cfm
% Please copy and paste the code instead of the example below. 
%

%\ccsdesc[300]{Computing methodologies~Learning from demonstrations}
%\ccsdesc[500]{Computing methodologies~Partially-observable Markov decision processes}

% \keywords{ACM proceedings, \LaTeX, text tagging}

\maketitle

%%%%%%%%%%%%%%%%%%%%%%%%%%%%%%%%%%%%%%%%%%%%%%%%%%%%%%%
\section{Introduction}
\label{sec:introduction}

Formula One (F1) is a class of motorsport, often described as its pinnacle, 
with the average annual cost of running a team in the hundreds of millions of pounds.\footnote{\href{https://www.forbes.com/sites/csylt/2020/01/14/red-bull-reveals-how-much-it-really-costs-to-run-an-f1-team/}{https://www.forbes.com/sites/csylt/2020/01/14/red-bull-reveals-how-much-it-really-costs-to-run-an-f1-team/}} Teams are thus in constant pursuit of the slightest gains in time, achieved by recruiting the best drivers and making their cars faster through ground-breaking engineering. Once a race begins, however, teams are unable to make changes to their cars or drivers but can optimise their selection of the \emph{tyre compounds} used through the race. Tyres \emph{degrade} and become slower as they are used and the different tyre compounds are selected to provide a trade-off between speed and rate of tyre degradation. 

This optimisation problem is the crux of \emph{race strategy}, which primarily consists of choosing which tyre compounds to select and when to make \emph{pitstops}, i.e. leaving the race to change tyres. The optimal race strategies are far from simple, however, given that live race situations must be taken into account, and thus have become a critical factor in determining the cars' finishing positions. This has been accentuated by F1's move to a single tyre manufacturer for all teams, rather than multiple competing tyre manufacturers, which has levelled the playing field between teams and increased the importance of their strategic decisions.\footnote{\href{https://www.autosport.com/f1/news/why-tyre-wars-have-largely-become-a-thing-of-the-past-in-motorsport/10526870/}{https://www.autosport.com/f1/news/why-tyre-wars-have-largely-become-a-thing-of-the-past-in-motorsport/10526870/}}

Currently, teams decide on candidate strategies before a race, doing their best to account for live situations which may occur such as \emph{safety cars}, i.e. periods where the cars must drive slower due to unsafe conditions on the track. They then run Monte Carlo simulations to test these different candidate strategies.\footnote{\href{https://www.oracle.com/customers/red-bull-racing-case-study/}{https://www.oracle.com/customers/red-bull-racing-case-study/}}\footnote{\href{https://www.mercedesamgf1.com/news/web-exclusive-q-and-a-with-rosie-wait}{https://www.mercedesamgf1.com/news/web-exclusive-q-and-a-with-rosie-wait}}\footnote{\href{https://www.autosport.com/f1/news/how-red-bull-comes-up-with-the-perfect-strategy-for-verstappen-and-perez/10619624/}{https://www.autosport.com/f1/news/how-red-bull-comes-up-with-the-perfect-strategy-for-verstappen-and-perez/10619624/}} By running millions of race simulations, teams are able to understand how well these strategies perform. However, this method is time-consuming, computationally expensive and mentally laborious since teams must first decide on the candidate strategies and then wait for simulation results. Additionally, since these strategies are pre-defined, they are unable to dynamically account for live race situations. Further, the methods for obtaining them do not readily explain their outputs.

In this work, we deploy reinforcement learning (RL) models in F1 race strategy, capitalising on their potential in real-time strategy applications, such as \emph{Go}~\cite{DBLP:journals/nature/SilverHMGSDSAPL16}, \emph{Atari} games~\cite{DBLP:journals/nature/MnihKSRVBGRFOPB15}, \emph{Starcraft}~\cite{DBLP:journals/nature/VinyalsBCMDCCPE19} and \emph{Dota 2}~\cite{Berner2019Dota2W}. We adapt and deploy an RL architecture for this task, demonstrating our approach's potential with a Monte Carlo race simulator of \DT{the Mercedes-AMG PETRONAS Formula One Team}. Further, we deploy explainable AI (XAI) techniques (see \cite{Ali_23} for a recent overview), to provide reasoning for the optimised strategies towards improving race strategists' trust in the models.

Concretely, we make the following contributions:
\begin{itemize}
    \item We introduce a flexible, extensible and portable architecture for real-time application of RL to F1 race strategy, allowing for training and deployment with different data sources; 
    \item We present \modelName{} (\modelNameFull{}), an RL model which outperforms baseline models in selecting race strategies for the 2023 Bahrain Grand Prix;
    \item In a generalisability study, we show how performance for one track or multiple tracks can be prioritised via training;
    \item We supplement \modelName{}   with three XAI techniques to explain its decisions, both in simulations and live races;
    \item We provide illustrations showing how \modelName{} replicates real-world strategic decisions made by strategists in live races. 
\end{itemize}

%%%%%%%%%%%%%%%%%%%%%%%%%%%%%%%%%%%%%%%%%%%%%%%%%%%%%%%

%%%%%%%%%%%%%%%%%%%%%%%%%%%%%%%%%%%%%%%%%%%%%%%%%%%%%%%
\section{Background and Related Work}
\label{sec:background}

\hspace*{3mm}
\textbf{RL.}
%\label{subsec:reinforcement_learning}
First, we give a brief overview of the RL model used in the paper.
\emph{Q-learning} is an off-policy Temporal Difference (TD) control method that uses a behavioural policy to control the exploration of the environment, whilst the target policy is updated. Actions are selected based on their respective \emph{Q-values} – the value of taking the action in the current state. Meanwhile, a \emph{deep Q-network} (DQN)~\cite{Mnih2013PlayingAW} is an RL agent that combines Q-Learning with a deep convolutional neural network. A DQN works by minimising the TD error, given the weights of the DQN. Finally, the deep recurrent Q-network (DRQN)~\cite{DBLP:conf/aaaifs/HausknechtS15}, builds upon the DQN by utilising a recurrent neural network to predict the Q-values instead of a regular convolutional neural network. In doing so, the model utilises previous states in its understanding and predictions of the current state, unlike the DQN which only considers the current state. The DRQN was found to parallel the performance of the traditional DQN and outperforms it in the Atari games \emph{Frostbite} and \emph{Double Dunk}~\cite{DBLP:conf/aaaifs/HausknechtS15}.

\textbf{RL in Motorsport.}
Many motorsport categories other than F1, such as Formula E, WEC, Indycar and GT Racing, also require effective race strategies. Within Formula E, Liu et al.~\cite{DBLP:journals/nca/LiuF20, DBLP:journals/kbs/LiuFA21} explore race strategy using \emph{distributed deep deterministic policy gradients}, neural networks and \emph{Monte Carlo Tree Search}.

In GT Racing, Boettinger and Klotz~\cite{DBLP:journals/corr/abs-2306-16088} implement a DQN to optimise strategy decisions over a limited action space. They find that a baseline Q-Learning model is significantly beaten by the DQN model. Meanwhile, Wurman et al.~\cite{Wurman2022OutracingCG} employ a \emph{quantile regression soft actor-critic} model to drive GT cars in the \emph{Gran Turismo} game. Their model outperformed the world's top drivers and demonstrated driving strategies such as different corner overtakes, highlighting its ability to learn human behaviours.

In F1, Heilmeier et al.~\cite{Heilmeier2020VirtualSE} introduce a \emph{virtual strategy engineer} comprising two separate neural networks that predict whether or not a pitstop is taken based on given race inputs. They discuss as future work the possibility of taking an RL-based approach, as we introduce in this paper.

Finally, Heine and Thraves~\cite{DBLP:journals/cejor/HeineT23} explore race strategy through dynamic programming (DP) and later incorporate random events such as safety car periods through stochastic DP. They find that both models can be solved to optimality. The DP approach allows the answering of questions such as ``If there is a safety car period in the current lap, is it worth making a pit stop?" and ``If yes, which tyre compound should we change to?". Comparatively, the stochastic DP approach considers delaying pitstops in favour of benefiting from future safety car periods, which improves upon the DP model when randomness is increased during simulations.

None of the works mentioned here, in F1 or otherwise, apply RL, coupled with explainability techniques, to optimise race strategy. 

\textbf{XAI for RL.}
XAI has gained prominence due to the growing role of AI in society, and thus, it is increasingly important that humans comprehend the predictions made by these systems. Milani et al.~\cite{Milani2024} give a useful review of different explainability techniques and present a novel taxonomy. Methods such as \emph{DAGGER}~\cite{DBLP:journals/jmlr/RossGB11}, \emph{VIPER}~\cite{DBLP:conf/nips/BastaniPS18}, and \emph{PIRL}~\cite{pmlr-v80-verma18a} convert RL policies into interpretable formats such as decision trees, whereas reward decomposition~\cite{Seijen_17} is used to decompose the rewards into a set of additive terms with semantic meaning. Some methods directly generate explanations. For example, natural language explanations employed by Hayes and Shah~\cite{DBLP:conf/hri/HayesS17} utilise templates for the agent to fill in. \emph{Saliency maps} are also used by Greydanus et al.~\cite{pmlr-v80-greydanus18a}, though they have been considered insufficient for explaining RL due to the subjective conclusions that can be drawn from their explanations~\cite{Milani2024}.

As for the methods we use in this paper, we first note that \emph{SHAP}~\cite{DBLP:conf/nips/LundbergL17} is a method to explain individual predictions based on \emph{Shapley values}~\cite{shapley:book1952}. SHAP explains the prediction of an instance by computing the contribution of each feature to the prediction using computed Shapley values. \emph{TimeSHAP} was then introduced by Bento et al.~\cite{DBLP:conf/kdd/0002SCFB21} and extends the use of Shapley values into sequential domains. TimeSHAP is a model-agnostic recurrent explainer that can compute feature-level attributions, highlighting the importance of features of the current time step to a prediction. 

Another method we use is \emph{VIPER}~\cite{DBLP:conf/nips/BastaniPS18}, an \emph{imitation learning} method that builds upon the data aggregation method of DAGGER~\cite{DBLP:journals/jmlr/RossGB11}. As a surrogate model, VIPER builds decision trees that closely mimic the behaviour of a pre-trained oracle. VIPER leverages the fact that the cumulative reward of each state-action pair is provided along with the optimal action, thus producing decision tree policies that are an order of magnitude smaller than those learned by DAGGER. 

Finally, Carreira-Perpiñán and Hada~\cite{DBLP:conf/aaai/Carreira-Perpinan21} generate counterfactuals for decision trees by constructing a linear programming problem, selecting the leaf node with the least distance to the current data point. For the distance function, the $\ell_2$ distance can be used to encourage changing all features, whilst the $\ell_1$ distance encourages the fewest number of features to be changed.

To exhibit a broad range of XAI approaches in this setting, we use TimeSHAP, VIPER and decision tree counterfactuals in this paper, and leave other deployments of XAI to future work.

%%%%%%%%%%%%%%%%%%%%%%%%%%%%%%%%%%%%%%%%%%%%%%%%%%%%%%%

%%%%%%%%%%%%%%%%%%%%%%%%%%%%%%%%%%%%%%%%%%%%%%%%%%%%%%%
\section{\modelNameFull{}}
\label{sec:main}
We now cover the architecture and implementation of our approach.

\subsection{Problem Formalisation}
% \fb{this problem definition seems quite long. More than problem definition, I'd say this is the "formalisation" of the problem.}

\hspace*{3mm}
\textbf{State Space.}
Selecting the state space, $\mathcal{S}$, is critical to ensure that the model can execute successful strategies without being overloaded with features. The chosen features are outlined in Table \ref{tab:features}. It should be noted that there are three tyre compounds used: soft, medium and hard. The soft tyres are initially the fastest but they wear down and become slower more quickly than the other tyres. On the other hand, the hard tyres are longer-lasting but slow, whilst the medium tyres give a compromise between the two. Meanwhile, the deployment of RL introduces a new challenge in data preprocessing since we do not have a predefined training dataset that dictates the limits of each feature. Theoretically, some features have no limit in their values. For instance, the gap to the leader $s_{gl}$, i.e. the time in seconds between the leader of the race and \modelName{}'s car, can be arbitrarily large if the driver takes a pitstop every lap, subsequently making them fall further and further back from the leader. %With different tracks and car paces, we cannot know the limit of the $s_{gl}$. 
We thus created a dataset with value ranges determined by observations from a reasonable number of simulations to estimate their working ranges. From this set of values, we created custom scaling functions that linearly scale each feature. %\DT{Removed the below}
% For example, we derive the scaling of the gap to leader feature, $s_{sgl}$ with Eq.~\ref{eq:scaling}. \todo{why these values? I think this raises more questions than answers. I would have it as before}

% \begin{equation}
%     s_{sgl} = \begin{cases}
%         -1 & \text{if } s_{gl} < -30 \\
%         1  & \text{if } s_{gl} > 200 \\
%         \frac{s_{gl}}{115} - \frac{17}{23} & \text{otherwise}
%     \end{cases}
%     \label{eq:scaling}
% \end{equation}

\begin{table*}[]
    \centering
    \small
    \begin{tabular}{llll}
        \hline
         \textbf{State Space, $\mathcal{S}$}                   & \textbf{Type}     & \textbf{Possible Values}  & \textbf{Description} \\ \hline
         *Terminal ($s_{terminal}$)         & Boolean           & True, False   & Whether or not the state is terminal.         \\
         Track ($s_{track}$)                & Track             & Enum of all tracks & The track on which the race is taking place.       \\
         Safety Car ($s_{sc}$)              & SafetyCarStatus   & Full, Virtual, None   & Whether or not the safety car is deployed, and to what level.      \\
         *Position ($s_{pos}$)              & Ordinal           & [1, ..., 20]  & The current position of the car being controlled. \\
         Scaled Position ($s_{spos}$)       & Continuous        & [0, 1]    & Linear scaling and clipping of $s_{pos}$.\\
         Race Progress ($s_{rp}$)           & Continuous        & [0, 1]    & The current percentage of the race completed. \\
         Current Tyre ($s_{tyre}$)          & TyreCompound      & Soft, Medium, Hard    & The current tyre compound fitted to the car being controlled. \\
        *Tyre Degradation ($s_{td}$)        & Continuous        & [?, ?]    & The time loss per lap due to the wear on the tyres for the car being controlled. \\
         Scaled Tyre Degradation ($s_{std}$)& Continuous        & [0, 1]    & Linear scaling and clipping of $s_{td}$ \\
         Soft Available ($s_{soft}$)        & Boolean           & True, False   & Whether or not the soft tyre is available to pit onto. \\
         Medium Available ($s_{medium}$)    & Boolean           & True, False   & Whether or not the medium tyre is available to pit onto. \\
         Hard Available ($s_{hard}$)        & Boolean           & True, False   & Whether or not the hard tyre is available to pit onto. \\
         *Gap Ahead ($s_{ga}$)             & Continuous        & [?, ?] & How many seconds the car positionally ahead of the car being controlled is.\footnotemark\\
         Scaled Gap Ahead ($s_{sga}$)       & Continuous        & [-1, 1]   & Linear scaling and clipping of $s_{ga}$ \\
        *Gap Behind ($s_{gb}$)              & Continuous        & [?, ?]    &  How many seconds the car positionally behind the car being controlled is. \\
         Scaled Gap Behind ($s_{sgb}$)      & Continuous        & [-1, 1]   & Linear scaling and clipping of $s_{gb}$\\
         *Gap to Leader ($s_{gl}$)           & Continuous        & [?, ?]   & How many seconds the car in the lead is ahead of the car being controlled.\\
         Scaled Gap to Leader ($s_{sgl}$)   & Continuous        & [-1, 1]   & Linear scaling and clipping of $s_{gl}$\\
         Last Lap to Reference ($s_{llr}$)  & Continuous        & [0, 2]    & Ratio of the last lap time to a reference lap time.\\
         Valid Finish ($s_{vf}$)            & Boolean           & True, False   & Whether or not the car being controlled can currently finish the race.\\
         \hline
    \end{tabular}
    \caption{The variables used in the reward function and the state space provided to the RL model, after data normalisation. One-hot encoding is used for all Enums and Boolean types. Data normalisation was applied by generating a dataset of likely values, running many simulations and applying linear scaling with clipping. Features labelled with * are not provided as an input to the model, but are used in the reward function.}
    \label{tab:features}
\end{table*}
\footnotetext{The car physically in front on the track may not be the car ahead in the positions due to overlapping. For example, a car may be in P20 ahead of another in P1 which is about to overlap it. This also applies to cars behind if they have been overlapped.}

\textbf{Action Space.}
At each lap, %the model 
\AR{\modelName{}} decides whether to change the tyres by executing a pitstop. Subsequently, the action space at any timestep, $t$, consists of four actions, $a^t \in \{\mathrm{\textit{no pit}},$ $\mathrm{\textit{pit soft}},$ $\mathrm{\textit{pit medium}},$ $\mathrm{\textit{pit hard}}\}$, representing taking no pitstop or taking a pitstop to one of the three tyre compounds\AR{, respectively}. The tyre that the model begins the race on is chosen by the Monte Carlo race simulator. Furthermore, the number of times each action can be selected during a race simulation is limited, since the available number of each tyre compound \AR{is specified in the F1 regulations}. 
%(a reflection of the regulations of F1).

\textbf{Reward Function.}
Rewards of $1$ are given for normal steps where the model progresses to the next lap. This can be done by not pitting, or by taking the first valid pitstop (i.e. a pitstop to a different tyre, a requirement in F1). A penalty of $-10$ is given for extraneous pitstops beyond the first valid pitstop. A terminal reward is also given based on the final finishing position which equates to 100 times the points the driver would receive in F1 for finishing in that position, where the points for P1 to P10 are 25, 18, 15, 12, 10, 8, 6, 4, 2, 1, respectively, and 0 for P11 to P20. For example, a P1 finish rewards 2500 points. By giving a reward of $1$ for normal steps, the model maximises its total reward by ensuring it steps through to the end. When the model fails a simulation, such as taking a pitstop to a tyre that is not available, it is penalised heavily with $-1000$, as in \cite{DBLP:journals/corr/abs-2306-16088}. Reward shaping is not used in the reward function because it is inherently difficult in this context. This is because we are unable to determine midway through a race whether a decision was good or not. Instead, we only know if a strategy was good at the end of a race when we discover the final finishing position. The reward function at any timestep $t$, $R^t$, is formally defined below, where $P = [25, 18, 15, 12, 10, 8, 6, 4, 2, 1]$: 
\begin{align*}
    R^t = \begin{cases}
        -1000 & \text{if } a^t = \text{\textit{pit soft}} \land \neg s^t_{soft} \\
        -1000 & \text{if } a^t = \text{\textit{pit medium}} \land \neg s^t_{medium} \\
        -1000 & \text{if } a^t = \text{\textit{pit hard}} \land \neg s^t_{hard} \\
        -1000 & \text{if } s^t_{terminal} \land \neg s^t_{vf} \\
        -10 & \text{if } a^t \neq \text{\textit{no pit}} \land s^t_{vf} \\
        100 * P[s^t_{pos} - 1] & \text{if } s^t_{pos} \leq 10 \land s^t_{terminal}\\
        0 & \text{if } s^t_{pos} > 10 \land s^t_{terminal}\\
        1 & \text{otherwise} \\
    \end{cases}
\end{align*}

In developing \modelName{}, we considered various architectures. The original DQN was motivated by Boettinger and Klotz's~\cite{DBLP:journals/corr/abs-2306-16088} approach. However, this model does not effectively capture the nature of partial observability whereby the agent only receives partial observations of the underlying state space. Consequently, we decided to utilise the DRQN architecture as outlined by Hausknecht and Stone~\cite{DBLP:conf/aaaifs/HausknechtS15}. %in the final model. 
%\todo{if we had better results with this over the DQN we can say that - We do, we just don't discuss the other models: (DQN gave HAM 2023 strategy, but achieved P5.3 on average compared to DRQN with 1-stop but P4.90 on average? - this was from Hyperparameter testing that we did (but we do not discuss it) }
We motivate this architecture through its ability to capture temporal dynamics which are important in this setting. For example, the model needs to understand whether it is catching the car ahead or falling behind it. This same issue %contrastingly 
motivated Heilmeier et al.~\cite{Heilmeier2020VirtualSE} to introduce new features representing these dynamics.

\subsection{System Architecture}
\label{subsec:system_architecture}

\begin{figure*}[ht]
    \centering
    \includesvg[width=0.88\textwidth]{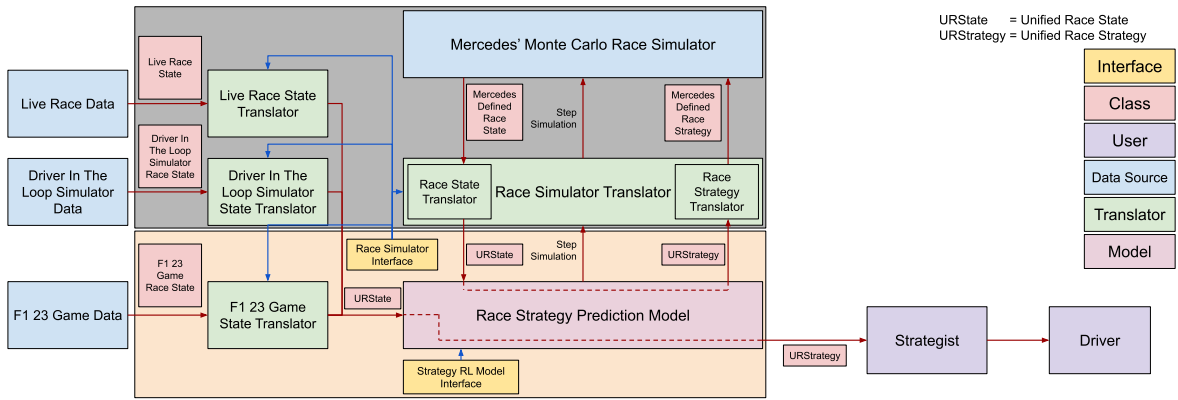}
    \caption{The system architecture implemented utilises abstraction to provide future flexibility, extensibility and portability. It allows for the substitution of different data sources for model prediction and the modification of race states and strategies% should their definitions be refined in future work
    .}
    \label{fig:system_architecture}
\end{figure*}

The design of the system architecture lends itself well to many features and is illustrated in Figure \ref{fig:system_architecture}. Central to the training loop is the black-box Monte Carlo race simulator. Since it handles proprietary data definitions, it is important to abstract our state and action space away from these definitions. By defining a custom \texttt{UnifiedRaceState} and \texttt{UnifiedRaceStrategy} classes with a translator layer in between, the model maintains its interoperability with the race simulator \DT{and decouples \AR{\modelName{}} from \AR{the black-box}}. Furthermore, the translator layer improves portability, allowing for the use of different data sources such as live race data \AR{or the \emph{F1 23} game.} %and enabling the use of the model during a live race. 
By creating a similar translator layer to convert live race data into the \texttt{UnifiedRaceState}, the model can be trained using the Monte Carlo race simulator, but deployed during a race. Such a method is currently not employed in industry, which defaults to the hard-coded race strategies. Portability with different data sources also enables the ability to refine the model using different information. For example, the model can be fine-tuned using the Driver-In-The-Loop Simulator to produce feasible strategies depending on a driver's racing style. The definition of the \texttt{UnifiedRaceStrategy} class also allows us to flexibly change the definition of a `race strategy' in the future, without significantly changing existing components. This could include the addition of other variables, e.g. target lap times or engine modes%, or lift-and-coast levels
. Implementing the RL model using an interface, similarly, allows the substitution of different model architectures and encourages the exploration of different approaches to find the best solution for each application of our approach.
%%%%%%%%%%%%%%%%%%%%%%%%%%%%%%%%%%%%%%%%%%%%%%%%%%%%%%%

%%%%%%%%%%%%%%%%%%%%%%%%%%%%%%%%%%%%%%%%%%%%%%%%%%%%%%%
\section{Evaluation}
\label{sec:evaluation}
In this section, we evaluate the performance of \modelName{}, analyse how well it generalises and examine the fidelity and comprehensibility of the implemented XAI techniques.
%\fb{let's find a name for the model} \todo{I'm not sure what to call it, any suggestions? AR: we have a week to think of something nice haha, DT: Could call it RSRL for Race Strategy RL, but pronounced like Russell (ironic name, and an easter egg for Mercedes :) - doesn't capture XAI in the name though)}

\subsection{Model Performance}
\label{subsec:model_performance}
To test the performance of a trained \modelName{} model, we take random seeds and run identical Monte Carlo simulations for \modelName{} and two baselines, namely a Fixed Strategy baseline and the industry state-of-the-art (SOTA) model \DT{provided by Mercedes}, and compare their results. To begin with, we start with one track for both training and testing:  Bahrain 2023 with a driver with a pace equivalent to an expected finish of P5.5 (i.e. in the third fastest team for Bahrain).\footnote{\AR{We leave assessing \modelName{} when trained on different drivers to future work. %However, due to limitations on simulation speed, training time is substantial and training many models is infeasible.
}} The Fixed Strategy model randomly employs one of the possible strategies defined on the F1 website for the 2023 Bahrain Grand Prix.\footnote{\href{https://www.formula1.com/en/latest/article/strategy-guide-what-are-the-possible-race-strategies-for-the-2023-bahrain.1nYrqBvGh8pZs9z6anDCdi}{https://www.formula1.com/en/latest/article/strategy-guide-what-are-the-possible-race-strategies-for-the-2023-bahrain.1nYrqBvGh8pZs9z6anDCdi}} \DT{Mercedes'} SOTA model \AR{is based on a probabilistic model using heuristics in its decision-making,} %and 
\AR{but is treated} %behaves 
as a black-box. %This model . 
%\todo{AR: we need some description of it that Stuart is okay with publishing - I'll check with him that what I've written is correct / acceptable, or get a different description}. 

\begin{figure}[t]
    \centering
    \includegraphics[width=\linewidth]{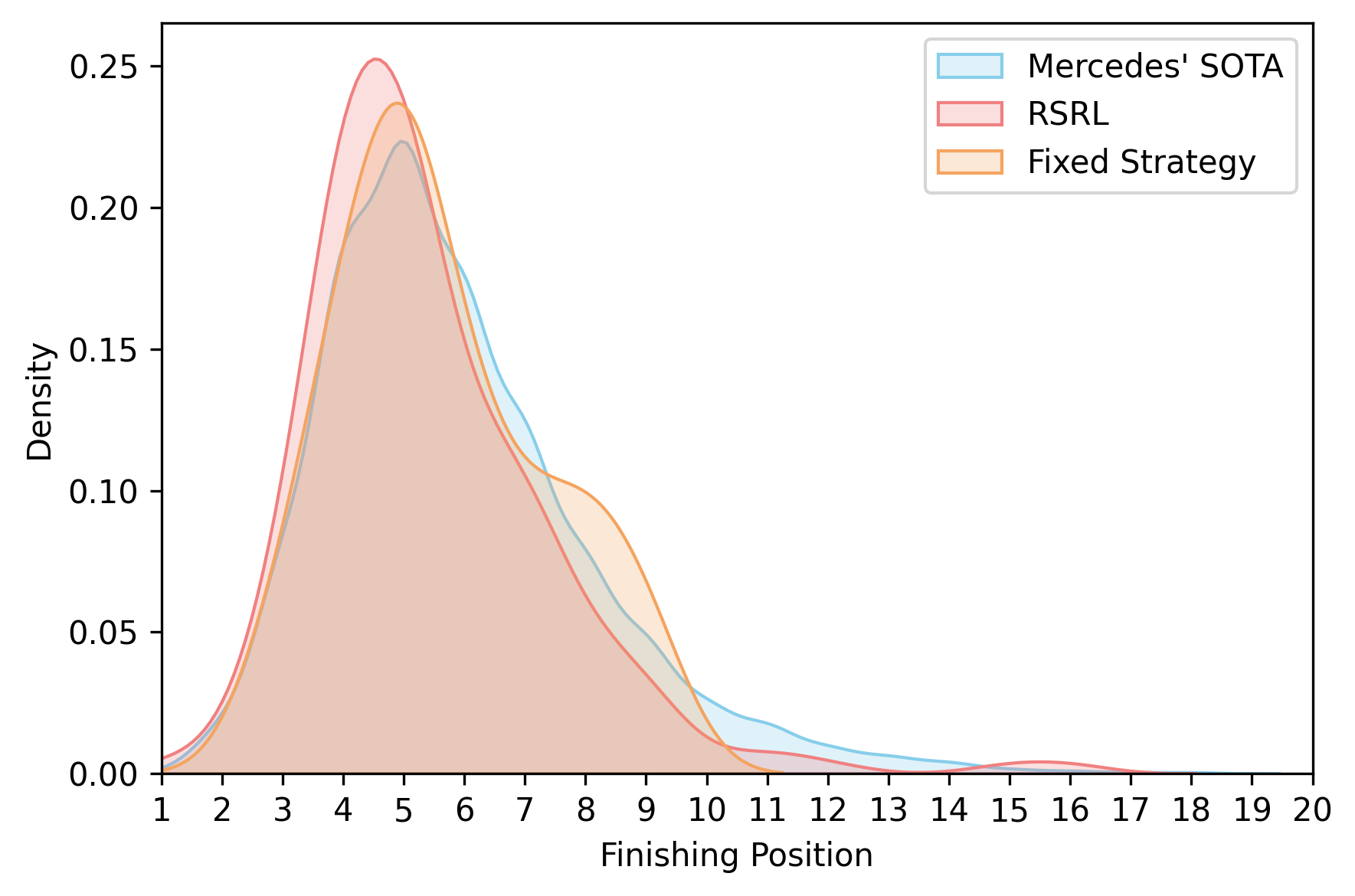}
    \caption{Average finishing positions for \DT{Mercedes'} SOTA model, \modelName{}, and the Fixed Strategy model.}
    \label{fig:density_plot}
\end{figure}

Through testing, we found that the best-performing \AR{version} of \modelName{} for the 2023 Bahrain Grand Prix achieved an average finishing position of P5.33 over 1920 simulations, whilst the Fixed Strategy model and \DT{Mercedes'} SOTA model achieved P5.63 and P5.86 respectively. It is important to note that whilst this appears to be a small improvement, there is a limit to the performance due to the simulations. For example, it is impossible to consistently finish P1 with the stochasticity of the race simulator and the modelled performance of the car. Improving upon both the Fixed Strategy and SOTA models thus demonstrates a significant achievement. Figure~\ref{fig:density_plot} demonstrates the improved performance of \modelName{} over Mercedes' and the Fixed Strategy models. The final hyperparameters selected for \modelName{} are listed in Table~\ref{tab:drqn_hyperparameters}. 

\begin{table}[t]
    \centering
    \begin{tabular}{ll}
        \hline
        \textbf{Hyperparameter}            & \textbf{Final Value}  \\ \hline
        \textbf{Epsilon}                   & 1                     \\
        \textbf{Epsilon Decay}             & 0.999                 \\
        \textbf{Minimum Epsilon}           & 0.005                 \\ 
        \textbf{Gamma}                     & 0.99                  \\
        \textbf{Learning Rate}             & 0.001                 \\ 
        \textbf{Weight Decay}              & 0.001                 \\ 
        \textbf{Replay Buffer Size}        & 1000                  \\ 
        \textbf{Episodes to Update Target} & 100                   \\
    \end{tabular}
    \caption{Final hyperparameters chosen for \modelName{}.}
    \label{tab:drqn_hyperparameters}
\end{table}

\subsection{Generalisability}
\label{subsec:generalisability}
To understand how well \modelName{} performs when trained across a different number of tracks, we trained three separate instances on 1, 2 and 9 tracks, named \modelName{}-1, \modelName{}-2 and \modelName{}-9, respectively. The training tracks represent a variety of expected strategies and exhibit diverse track characteristics. To determine how well these models generalise, we tested each trained model on the tracks that they had and had not seen. Each model was tested approximately 900 times per track. Table \ref{tab:generalisability_table_1} shows the average finishing positions.

It can be seen that the models trained on a few tracks, \modelName{}-1 and \modelName{}-2, finish on average 0.93 and 1.77 positions ahead of the SOTA model respectively when tested on their training tracks. However, they finish 6.40 and 4.01 positions behind when tested on their unseen tracks. Comparatively, \modelName{}-9, finishes 0.87 positions behind the \DT{Mercedes'} SOTA model on its training tracks, but 0.21 positions ahead for the unseen tracks. These results demonstrate how models can be trained to complete multiple races, improving the generalisability of race strategies, but at the cost of performance when they are trained in the same time frame.

\begin{table*}[ht]
    \centering
    \begin{tabular}{cccccccccccccccc}
        \cline{2-16}
         \textbf{Models} & \textbf{JPN} & \textbf{BHR} & \textbf{AZE} & \textbf{GBR} & \textbf{HUN} & \textbf{ITA} & \textbf{SGP} & \textbf{QAT} & \textbf{ABU} & \textbf{SPN} & \textbf{SAU} & \textbf{AUT} & \textbf{MEX} & \textbf{USA} & \textbf{Average} \\ 
         \hline
         % \textbf{1 Track RL} & 
         \textbf{\modelName{}-1} & 
         \underline{4.45} & 
         9.62 & 
         12.33 & 
         6.09 & 
         12.29 & 
         16.25 & 
         5.29 & 
         8.12 & 
         15.75 & 
         5.57 & 
         6.77 & 
         8.29 & 
         8.33 &
         6.75 &
         8.99 \\
         %
         % \textbf{2 Track RL} &  IF REVERTING, MAKE SURE TO REMOVE FROM THE OTHER PLACES IN THE TEXT
         \textbf{\modelName{}-2} & 
         \textbf{\underline{3.75}} & 
         \underline{6.47} & 
         15.33 & 
         15.67 & 
         8.83 & 
         13.50 & 
         4.33 & 
         18.00 & 
         13.75 &
         \textbf{2.50} &
         9.25 &
         13.60 &
         13.50 &
         12.00 &
         10.75 \\
         %
         % \textbf{9 Track RL} & 
         \textbf{\modelName{}-9} & 
         \underline{4.94} & 
         \underline{6.45} & 
         \underline{7.36} & 
         \underline{6.07} & 
         \underline{\textbf{4.23}} & 
         \underline{10.96} & 
         \underline{4.02} & 
         \underline{3.96} & 
         \underline{11.26} &
         3.06 & 
         6.83 & 
         6.06 & 
         4.36 & 
         2.71 &
         5.88 \\
         \hline
         \!\!\textbf{Fixed Strategy}\!\! & 
         6.42 & 
         \textbf{5.63} & 
         9.67 & 
         7.07 & 
         6.05 & 
         \textbf{7.96} & 
         \textbf{3.60} & 
         6.63 & 
         7.95 & 
         6.65 & 
         \textbf{6.33} & 
         \textbf{6.02} & 
         6.07 & 
         \textbf{2.22} &
         6.31 \\ 
         \hline
         \!\!\!\textbf{\DT{Mercedes} \AR{SOTA}}\!\!\! & 
         6.22 & 
         5.86 & 
         \textbf{6.95} & 
         \textbf{5.00} & 
         4.33 & 
         9.01 & 
         3.88 & 
         \textbf{2.93} & 
         \textbf{7.25} &
         2.87 & 
         6.44 & 
         7.77 & 
         \textbf{3.96} & 
         3.02 &
         5.39 \\ 
         \hline
    \end{tabular}
    \caption{Average finishing position for each model when tested across 14 tracks. These tracks are: \underline{J}A\underline{P}A\underline{N}, \underline{B}A\underline{HR}AIN, \underline{AZE}RBAIJAN, \underline{G}REAT \underline{BR}ITAIN, \underline{HUN}GARY, \underline{ITA}LY, \underline{S}IN\underline{G}A\underline{P}ORE, \underline{QAT}AR, \underline{ABU} DHABI, \underline{SP}AI\underline{N}, \underline{SAU}DI ARABIA, \underline{AU}S\underline{T}RIA, \underline{MEX}ICO, and \underline{USA}. The underlined font represents tracks in each of the RL models' training data, and the bold font highlights the best-performing model for each track.}
    \label{tab:generalisability_table_1}
\end{table*}

As a qualitative assessment% to complement the aforementioned quantitative analysis
, we analyse the resulting tyre strategies of the three models and compare them with the tyre strategies taken by the \DT{Mercedes'} SOTA model and two of the pre-defined strategies employed by the Fixed Strategy model. Strategies are written in a shortened form where S[10, 20]M indicates a Soft-Medium strategy with the pitstop occurring between laps 10 and 20. %From this analysis, 
\AR{In this analysis, we look for reasonable strategies, i.e. those with:}
%. We define reasonable strategies as those with:

\begin{itemize}
    \item between one and three pitstops;
    \item a harder compound being used for more laps than a softer compound; and
    \item visual similarity to or a better average finishing position (listed in Table~\ref{tab:generalisability_table_1}) than \DT{Mercedes'} SOTA or the Fixed Strategy models.
\end{itemize}

\begin{figure*}
    \centering
    \begin{minipage}{.49\textwidth}
      \centering
      \includegraphics[width=\linewidth]{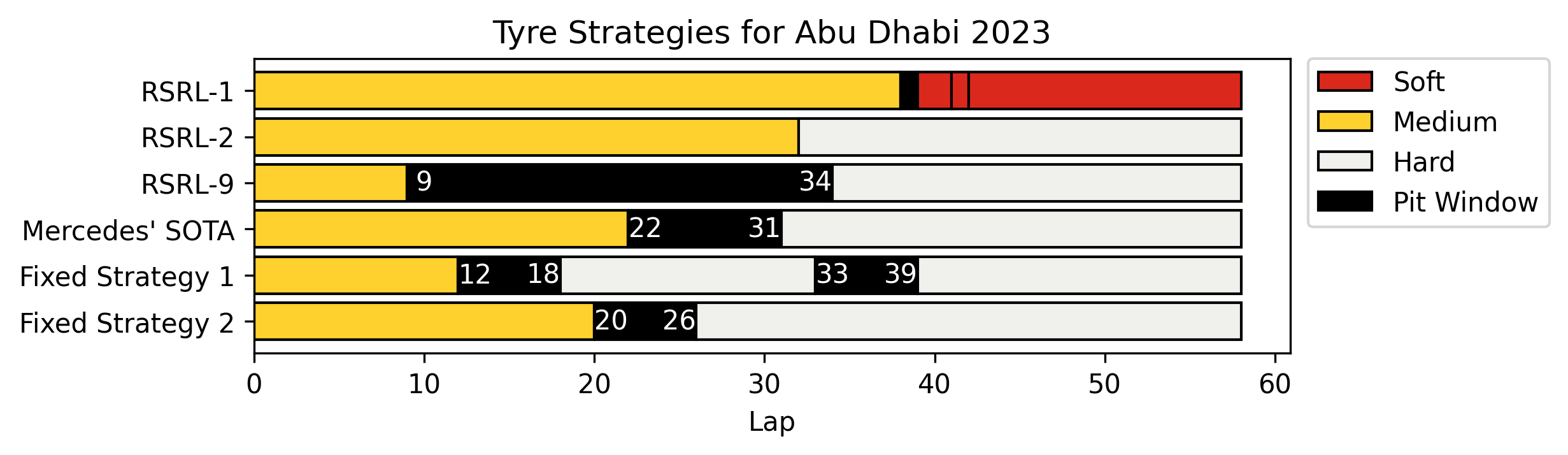}
    \end{minipage}%
    \begin{minipage}{.49\textwidth}
      \centering
      \includegraphics[width=\linewidth]{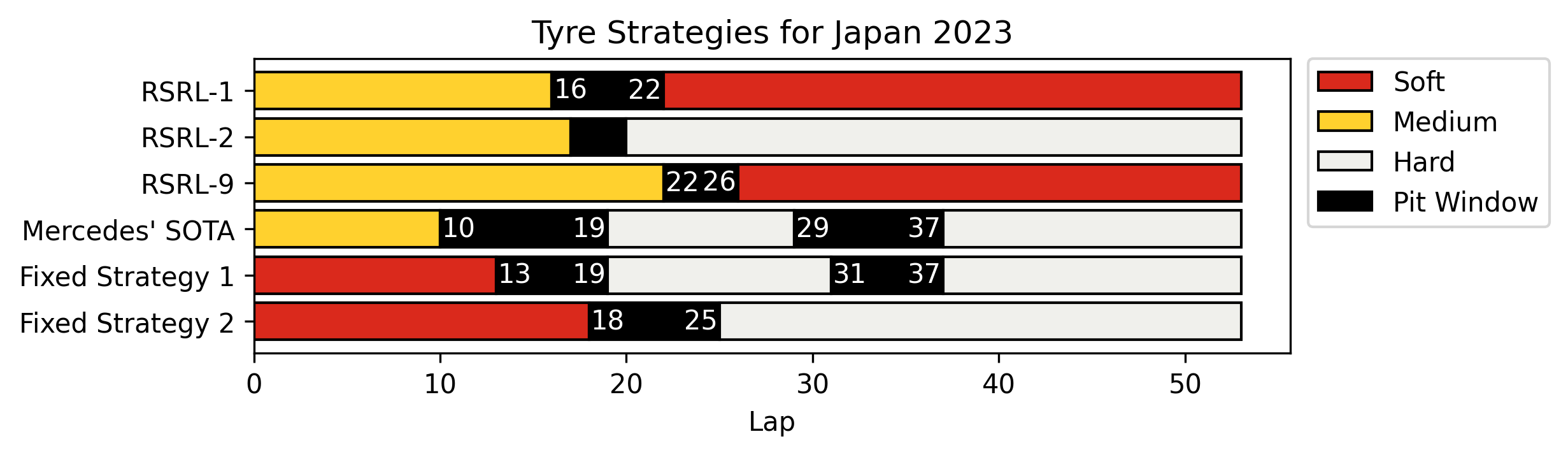}
    \end{minipage}
    \begin{minipage}{.49\textwidth}
      \centering
      \includegraphics[width=\linewidth]{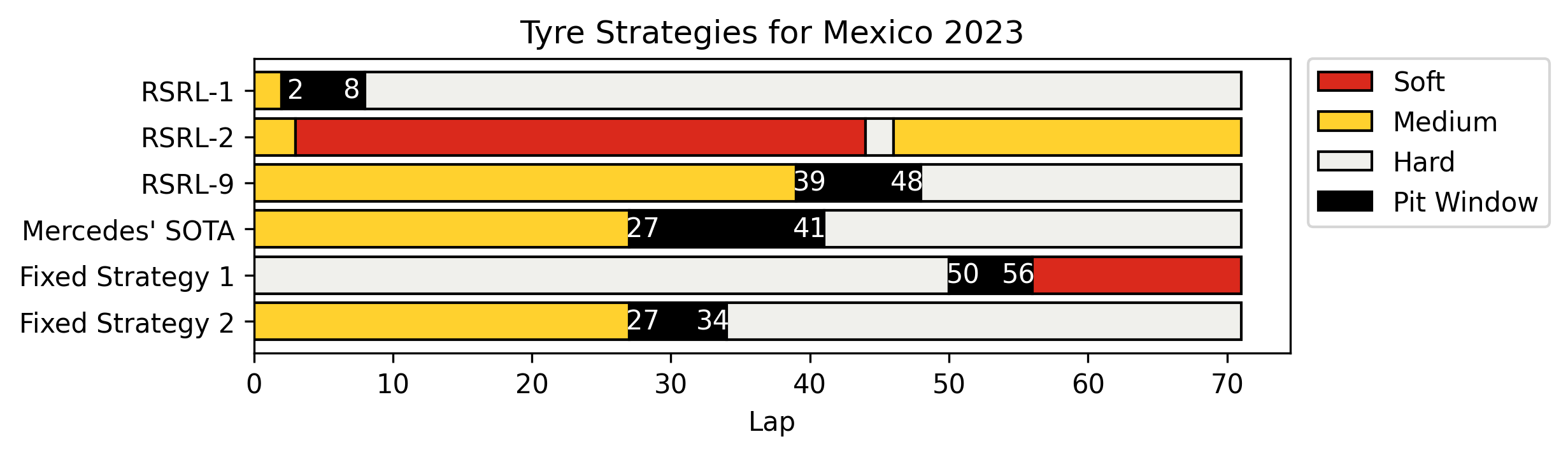}
    \end{minipage}
    \begin{minipage}{.49\textwidth}
      \centering
      \includegraphics[width=\linewidth]{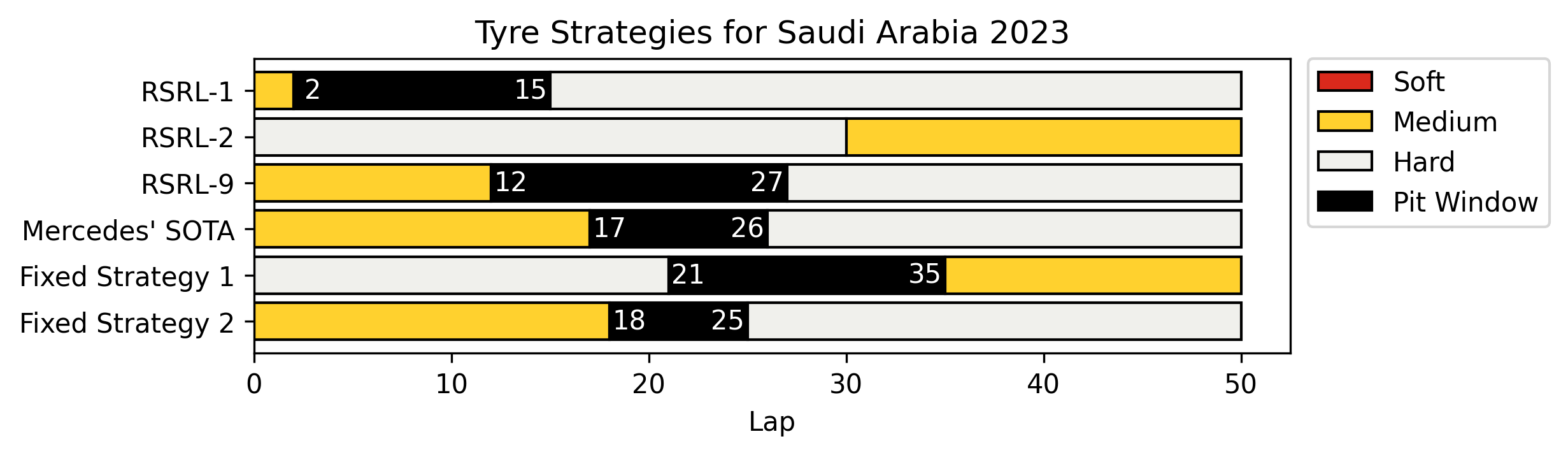}
    \end{minipage}
    
    \caption{The most common tyre strategies generated by each generalisability model, \DT{Mercedes'} SOTA model, and two of the fixed strategies from the Fixed Strategy model for the Abu Dhabi, Japan, Mexico and Saudi Arabia Grands Prix. The black `Pit Window' represents an average period in which a pitstop is executed.
    % \todo{also, are these the most popular chosen strategies for each model? or do they always select these? I thought the simulator puts them on random tyres? - DT: these are the most common strategies chosen. The tyres are not quite random, they're biased towards some compound tyre usually, but it can choose another.}}
    \label{fig:gen_ts}}
\end{figure*}

\textbf{1-Track Model.%: \modelName{}-1
}
For \modelName{}-1, we note how the strategy taken for Japan (its only training track) is a relatively reasonable M[16,22]S strategy, seen in Figure~\ref{fig:gen_ts} though the soft tyre is used for longer than the medium. However, the model successfully employs a one-stop strategy, similar to Fixed Strategy 2. From Table~\ref{tab:generalisability_table_1}, we see how this strategy outperforms the Fixed Strategy model, finishing an average of 1.97 positions ahead. The model struggles with the Abu Dhabi and Mexico tracks, where the strategies chosen are not reasonable. For instance, in Abu Dhabi, seen in Figure~\ref{fig:gen_ts}, the chosen strategy is M[38,39]S[41]S[42]S, which finishes on average 8.5 positions behind \DT{Mercedes'} SOTA model.

\textbf{2-Track Model.%: \modelName{}-2
}
\modelName{}-2 similarly demonstrates reasonable strategies for its training tracks, Japan and Bahrain. The M[17,20]H strategy from Japan reflects the S[18,25]H strategy from the Fixed Strategy 2 model. Analogously to \modelName{}-1, it produces mixed results on the unseen tracks, with unreasonable strategies being used in the Mexico race, i.e. M[3]S[44]H[46]M, as shown in Figure~\ref{fig:gen_ts}. Compared to the other models' strategies, \modelName{}-2 takes too many pitstops, performs the worst of all the models, and finishes on average 9.54 positions behind \DT{Mercedes'} SOTA model.

\textbf{9-Track Model.%: \modelName{}-9
}
\modelName{}-9 demonstrates reasonable strategies across all of the races. Notably, we see how strategies for the Abu Dhabi, Japan, Mexico and Saudi Arabia races all resemble strategies employed by \DT{Mercedes'} SOTA model and the Fixed Strategies. \AR{However, t}his model demonstrates a very wide window for its pitstop for the Abu Dhabi race\AR{. I}ts lack of precision results in its poorer average finishing position, being 4.01 positions behind \DT{Mercedes'} SOTA model, but still ahead of \modelName{}-1 and \modelName{}-2. \modelName{}-9 is the only model to outperform \DT{Mercedes'} SOTA model in Hungary, selecting a one-stop strategy compared to the \DT{Mercedes} SOTA model's two-stop, and finishes on average 0.1 positions ahead.

\subsection{Explanations}
To evaluate the efficacy of the explanations, we undertake (high-level) assessments of their \emph{fidelity}, i.e. how closely the explanations replicate the models, and \emph{comprehensibility}, i.e. how comprehensible the explanations are to users. 
To demonstrate the use of the explanations in practice, we also exemplify how each of the explanations highlights the motivations behind the decisions of \modelName{} (from Section~\ref{subsec:model_performance}) in a race simulation. To do so, we analyse \modelName{}'s decision of \emph{no pit} on lap 10 of the 2023 Bahrain Grand Prix.

\textbf{TimeSHAP Feature Importance.}
To determine the fidelity of the feature importance, we measure the Mean Absolute Error (MAE) for 100 timesteps across 10 simulations. \modelName{} achieved an average error of 124.39. Compared to the maximum reward of 2500, this represents a 5\% normalised MAE, demonstrating that the importance values are accurate and consistent with the actual rewards, thus reducing the likelihood of large errors when predicting the feature importance.

Feature importance is easy to comprehend, whereby larger bars are considered more important to a model's predictions. Even without explicit SHAP values, it is trivial to understand the importance of each feature to the other features~\cite{DBLP:journals/corr/abs-2309-11987}.

\begin{figure}[]
    \centering
    \includegraphics[width=\linewidth]{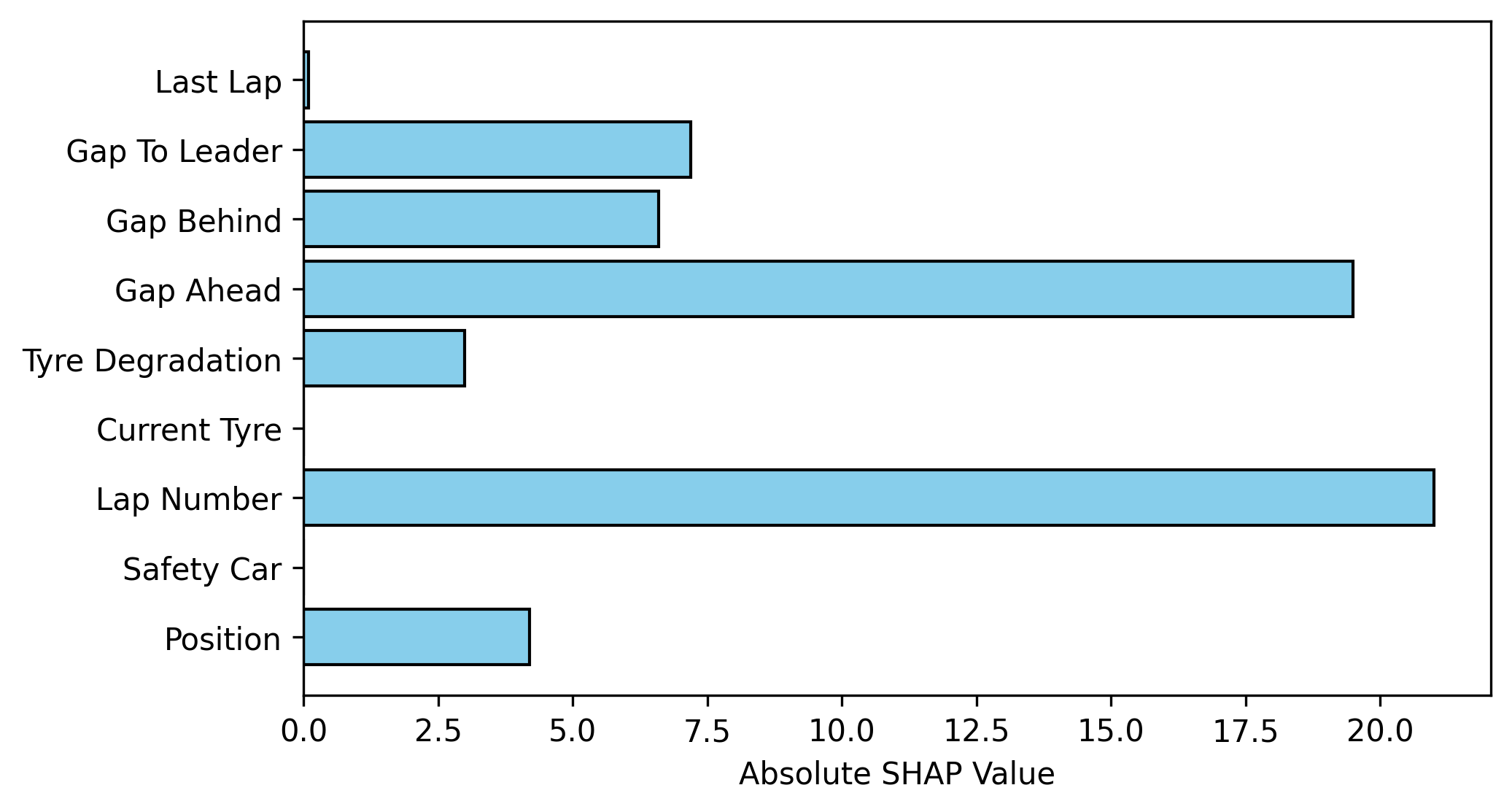}
    \caption{The feature importance for \modelName{} on lap 10 of the 2023 Bahrain Grand Prix.\protect\footnotemark}
    \label{fig:lap10_fi}
\end{figure}

A feature importance plot for lap 10 of the 2023 Bahrain Grand Prix is presented in Figure~\ref{fig:lap10_fi} indicating how \modelName{} is looking towards the car ahead, evidenced by the dominant Gap Ahead. The Lap Number plot is the most influential, indicating how \modelName{} is looking towards its expected pit window since observing the lap number informs \modelName{} of when to pit.

\footnotetext{These feature names have been reverted from their scaled counterparts that are passed into the model. For example, Gap Ahead is displayed, compared to Scaled Gap Ahead which is used by the model. These feature names have been swapped for clarity.}

\begin{figure}
    \centering  \includegraphics[width=0.68\linewidth]{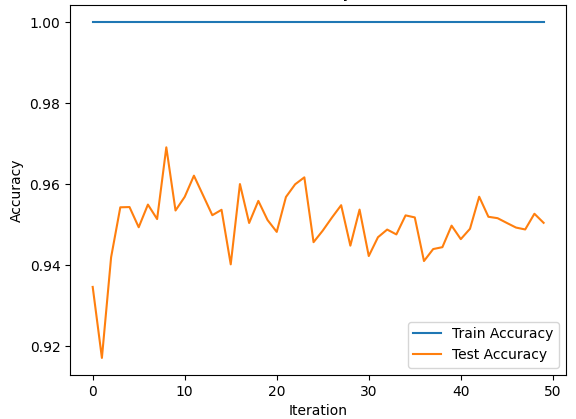}
    \caption{\DT{Accuracy during training and testing of 50 iterations of VIPER decision trees. Each iteration covers all previous data points queried from the oracle (the best \modelName{} model), up to 26,714 in iteration 50.
    }}
    \label{fig:viper_train}
\end{figure}

\textbf{VIPER Decision Tree Surrogate Model.}
The fidelity of the VIPER decision tree surrogate model is demonstrated with a confusion matrix generated from 100 random simulations, seen in Table~\ref{tab:dt_confusion_matrix}. We note that the accuracy is 0.926, however, there are considerably more \emph{no pit} decisions being taken through the 100 simulations. The F1 score of 0.910 represents a good balance between precision and recall, indicating that \modelName{} is good at identifying both positive and negative cases for each class. \DT{Furthermore, Figure~\ref{fig:viper_train} depicts the change in testing accuracy of the VIPER model over 50 iterations during training. A peak testing accuracy of 0.97 is achieved against the oracle, supporting the strong fidelity of the model.}
% Furthermore, Figure~\ref{fig:viper_train} demonstrates the testing accuracy of the final VIPER decision tree. By evaluating 50 students, we achieve a peak testing accuracy of 0.97 against the oracle, supporting our strong fidelity claim.

\begin{table}[]
    \centering
    \begin{tabular}{cccccc}
        \cline{3-6}
        & & \multicolumn{4}{c}{Predicted Label}\\
        & & \textbf{np} & \textbf{ps} & \textbf{pm} & \textbf{ph} \\ \hline
        \multirow{4}{3em}{True Label} & \textbf{np} & 4699 & 9 & 10 & 62 \\
        & \textbf{ps} & 55 & 0 & 1 & 2 \\
        & \textbf{pm} & 55 & 0 & 6 & 0 \\
        & \textbf{ph} & 180 & 4 & 0 & 39 \\
    \end{tabular}
    \caption{Confusion matrix for the VIPER-trained decision tree of the best-trained model for predicted labels: 
    \emph{\underline{n}o \underline{p}it}, 
    \emph{\underline{p}it \underline{s}oft}, \emph{\underline{p}it \underline{m}edium} and \emph{\underline{p}it \underline{h}ard}.}
    \label{tab:dt_confusion_matrix}
\end{table}

\begin{figure}
    \centering
    \includegraphics[width=0.68\linewidth]{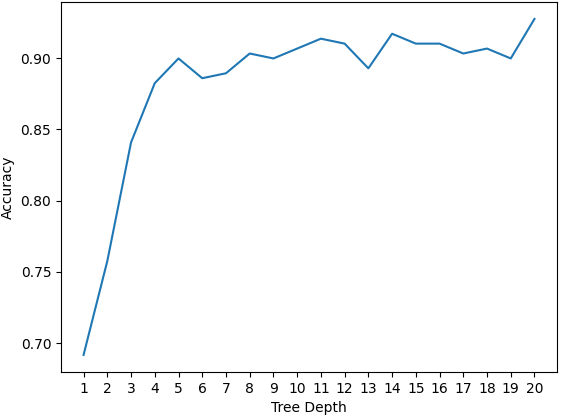}
    \caption{Accuracy of VIPER decision tree testing with different maximum node depths.}
    \label{fig:tree_depth_accuracy}
\end{figure}

Trained decision trees vary in comprehensibility depending on their depth. In \cite{DBLP:conf/nips/BastaniPS18}, the authors generate trees with hundreds of nodes which becomes very difficult to comprehend. From depth testing the tree, we found that a maximum depth of 4 to 6 maintains good accuracy and looks to be sufficiently comprehensible. Figure~\ref{fig:tree_depth_accuracy} demonstrates how we reach a plateau around this range. However, %it should be noted that 
with a larger input state space, this value is likely to increase as the tree has to split the dataset even further to achieve high accuracy.

Looking at the decision path for the \emph{no pit} decision on lap 10 of the 2023 Bahrain Grand Prix (Table \ref{tab:lap10_dt}), we see that predictions are influenced by the tyres, race progress and the gaps. Similarly to the feature importance, we see that the race progress being less than 26\% (15 laps) is an important factor. The first node decision on the current fitted tyre indicates how the model has different strategies based on the starting tyre, which is randomly chosen by the race simulator. The last lap to reference, $s_{llr}$ feature value being $\leq 1.016$ indicates that the decision tree is also looking at the lap times, making its decision based on whether or not the current tyres begin slowing down significantly.

\begin{table}[]
    \centering
    \begin{tabular}{lll}
         \textbf{Formal Definition}                 & \textbf{Natural Language Translation} \\ \hline
         $\neg(s_{tyre} = \text{Hard})$             & No Hard \\
         $s_{td} \leq 2.560$                        & *Tyre Degradation $\leq$ 2.560 \\
         $s_{rp} \leq 0.263$                        & Race Progress $\leq$ 0.263 \\
         $s_{gl} \leq 35.550$                       & *Gap To Leader $\leq$ 35.550 \\
         $s_{ga} > 4.560$                           & *Gap Ahead $>$ 4.560 \\
         $s_{gb} > 1.740$                           & *Gap Behind $>$ 1.740 \\
         $s_{gb} > 1.800$                           & *Gap Behind $>$ 1.800 \\
         $s_{llr} \leq 1.016$                       & Last Lap To Reference $\leq$ 1.016 \\
         $s_{gb} \leq 6.390$                        & *Gap Behind $\leq$ 6.390 \\
         $s_{td} > 0.860$                           & *Tyre Degradation $>$ 0.860 \\
         $s_{gb} > 0.300$                           & *Gap Behind $>$ 0.300 \\

    \end{tabular}
    \caption{Decision Tree Path of the VIPER decision tree for \modelName{} on Lap 10 of the 2023 Bahrain Grand Prix, resulting in its predicted \emph{no pit} decision. Features with * have had their scaling inversed for clarity.\protect\footnotemark}
    \label{tab:lap10_dt}
\end{table}
\footnotetext{These feature names have been reverted to their unscaled counterparts for clarity and a better understanding of the model's decisions. For example, Gap To Leader > 35.550 means that the decision at this node was taken based on whether or not the leader was more than 35.550 seconds ahead. %This is not easily inferable without inversing the scaling.
}

\textbf{Decision Tree Counterfactuals.}
To measure the fidelity of counterfactuals, we examine the proximity of generated counterfactuals to the original instance by counting the number of feature changes required to reach alternate decisions. By changing fewer features, the counterfactual demonstrates high fidelity because it suggests that the counterfactual closely aligns with the model's decision boundaries and feature importance. Taking 100 random simulations, and computing the closest counterfactual for a random alternate action, the average number of features requiring changing for the closest counterfactual was 1.630 whilst the average distance was 0.069. These low figures represent how these counterfactuals are sufficiently close to the real prediction, only requiring small but critical changes to result in a different action. This is in line with Molnar and Dandl's~\cite{DBLP:conf/ppsn/DandlMBB20} criteria for good counterfactuals, whereby they must be diverse and contain likely features.

Decision Tree Counterfactuals are presented in a textual format, listing exactly what changes are required to predict the alternate action. By demonstrating to the user exactly what changes are required, %there is little room for misunderstanding and a lack of communication, thus improving comprehensibility.
\AR{we posit that they are reasonably comprehensible to users.}

On lap 10 of the 2023 Bahrain Grand Prix, \AR{the explanations show} what it would take for us to take an early pitstop, \AR{i.e.} %e.g. 
converting to a strategy with an extra pitstop. Looking at a hypothetical \emph{pit soft} decision, the counterfactual tells us that we need to complete 4.503 more laps. Interestingly, this would take us to lap 14, the lap on which the \AR{(real-world) race winner} Max Verstappen pitted for soft tyres in the 2023 Bahrain Grand Prix. 

%%%%%%%%%%%%%%%%%%%%%%%%%%%%%%%%%%%%%%%%%%%%%%%%%%%%%%%

%%%%%%%%%%%%%%%%%%%%%%%%%%%%%%%%%%%%%%%%%%%%%%%%%%%%%%%
\section{Conclusions and Future Work}
\label{subsec:conclusions_and_future_work}
We have introduced \modelName{}, an RL model for F1 Race Strategy. \modelName{} achieves an average finishing position of P5.33, outperforming the Fixed Strategy Model's P5.63 and \DT{Mercedes'} SOTA model's P5.86 at the 2023 Bahrain Grand Prix. Generalisability testing highlighted how when more tracks are used in training, the model performs worse on seen tracks due to a lack of %dedicated training on each track and thus a lack of 
state space exploration, %. However, by training on multiple tracks, the model can generalise well and thus perform better when approaching unseen tracks. 
\AR{but better on unseen tracks.}
Finally, we concluded that the explainability methods support \modelName{}'s predictions and help to explain its decisions, as evident through our examples. Drawing parallels between the simulations and real-world phenomena with these explanations indicates how these models are deployable in practice. %This work marks the first application of explainable RL in motorsport, giving increased performance over the SOTA model, the replication of real-world phenomena and immediate strategy decision response time. 
This work also extends beyond %the realm of 
F1, providing a framework for implementing portable and extendable RL systems through an abstraction architecture and %marking the introduction of
deploying methods for explaining \modelName{}'s decisions in real-time strategy applications.

This work opens up numerous avenues for future work, not least controlling multiple cars to explore cooperative strategies presents an interesting venture towards discovering new strategies. Additionally, we \AR{would like to} to improve strategy predictions based on a driver's driving style by fine-tuning models to suggest %complementary 
\AR{tailored and personalised} strategies. Finally, given that we have shown what can be achieved in learning these strategies from scratch, we would like to improve this process by incorporating human-in-the-loop feedback processes \AR{to utilise race strategists' priceless expertise.}
%, allowing strategists and the RL models in tandem to push the limits of what can be achieved.

%%%%%%%%%%%%%%%%%%%%%%%%%%%%%%%%%%%%

%%%%%%%%%%%%%%%%%%%%%%%%%%%%%%%%%%%%%%%%%%%%%%%%%%%%%%%
\section*{Acknowledgements}

The research described in this paper was partially supported by the EPSRC (grant number EP/X015823/1).

\bibliographystyle{ACM-Reference-Format}
\bibliography{ref}

\end{document}